\documentclass[letterpaper, 10 pt, conference]{ieeeconf}  

\IEEEoverridecommandlockouts                              

\usepackage{graphics} 
\usepackage{epsfig}
\usepackage{mathptmx} 
\usepackage{times} 

\usepackage[ruled,vlined,linesnumbered]{algorithm2e}
\usepackage{amsmath,amssymb,amsfonts}
\usepackage{subcaption}
\usepackage{algorithmic}
\usepackage{color}
\usepackage[font={small},belowskip=0pt,aboveskip=3pt]{caption}
\usepackage{url}
\usepackage{balance}

\usepackage{multirow}
\usepackage{array}
\usepackage{booktabs}
\newcolumntype{L}[1]{>{\centering\arraybackslash}m{#1}}

\newtheorem{theorem}{Theorem}

\title{ \bf
LEA*: An A* Variant Algorithm with Improved Edge Efficiency \\
for Robot Motion Planning
}

\author{Dongliang Zheng$^{1}$ and Panagiotis Tsiotras$^{2}$
\thanks{This work has been supported by NSF award IIS-2008695.}
\thanks{$^{1}$Dongliang Zheng is with School of Aerospace Engineering,
        Georgia Institute of Technology, Atlanta, GA 30332, USA. Email:
        {\tt\small dzheng@gatech.edu}}%
\thanks{$^{2}$Panagiotis Tsiotras is with School of Aerospace Engineering and Institute for Robotics and Intelligent Machines, Georgia Institute of Technology, Atlanta, GA 30332, USA. Email:
        {\tt\small tsiotras@gatech.edu}}%
}

\begin{document}

\maketitle
\thispagestyle{empty}
\pagestyle{empty}

\begin{abstract}

In this work, we introduce a new graph search algorithm, lazy edged based A* (LEA*), for robot motion planning.
By using an edge queue and exploiting the idea of lazy search, LEA* is optimally vertex efficient similar to A*, and has improved edge efficiency compared to A*.
LEA* is simple and easy to implement with minimum modification to A*, resulting in a very small overhead compared to previous lazy search algorithms.
We also explore the effect of inflated heuristics, which results in the weighted LEA* (wLEA*). 
We show that the edge efficiency of wLEA* becomes close to LazySP and, thus is near-optimal. 
We test LEA* and wLEA* on 2D planning problems and planning of a 7-DOF manipulator.
We perform a thorough comparison with previous algorithms by considering sparse, medium, and cluttered random worlds and small, medium, and large graph sizes. 
Our results show that LEA* and wLEA* are the fastest algorithms to find the plan compared to previous algorithms. 

\end{abstract}

\section{Introduction}
We consider the shortest path problem on a graph having in mind path planning for robotics applications.
A graph $G$ is given by a vertex set $V$, an edge set $E$, and an edge cost/weight set $W$.
Given two query vertices $v_i, v_j \in V$, the shortest path problem is to find the minimum cost path $\tau^*$ connecting $v_i$ and $v_j$ through graph $G$, if such a path exists.

For robot path planning, sampling-based methods \cite{kavraki1996probabilistic, bohlin2000path, gammell2020batch} are popular methods to make the problem tractable. The planning space is abstracted as a graph, where vertices represent robot configurations and edges represent movements of the robot. Then, graph search algorithms are employed to find a path for the robot. 
Numerous algorithms have been developed to solve the shortest path problem. For example, A* \cite{Hart1968A} is a popular algorithm that is vertex optimal. That is, any other algorithm that finds the same shortest path will expand at least as many vertices when using the same heuristic. 

The procedure of computing the cost of an edge is called \textit{edge evaluation}.
One practical issue with robot path planning is that the edge costs $W$ may be unknown when starting the graph search. 
This is the case when the environment is unknown, and we do not know if the edges are in collision with obstacles or not. 
Then, edge evaluations are performed online as part of the graph search algorithm to compute the edge cost (e.g., perform collision checking).
Even if the environment is known before starting the graph search, evaluating all edges before starting the graph search is unnecessary since only some of the edges will probably be visited when searching for the solution. 
Thus, performing edge evaluation online will save time compared to evaluating all edges before the graph search.

In many robotic motion planning problems, edge evaluation, including performing collision checking, is the major computational bottleneck \cite{LaValle2006Planning}. 
Lazy search algorithms that intend to reduce the number of edge evaluations, have been developed. 
Two notable lazy search algorithms are LWA* \cite{Cohen2015Planning} and LazySP \cite{Dellin2016A}. 
Lazy search algorithms use a heuristic for the edge cost, which is easy to compute, and provides a lower bound of the true edge cost. 
The edge heuristic is used to guide the graph search so that edge evaluation is done only when necessary.
LazySP is also proved to be optimal with respect to minimizing edge evaluations \cite{Mandalika2018Lazy}.

Intuitively, Lazy search algorithms reduce the overall planning time if edge evaluations dominate the planning time by reducing the number of edge evaluations. 
However, lazy search algorithms reduce edge evaluations at the expense of extra graph operations (vertex expansion, calculating current best plan), which may introduce substantial computational overhead compared to A*.

The ultimate goal is to achieve optimal edge efficiency while at the same time, introduce minimum computational overhead. In this paper, we propose the LEA* algorithm, which is shown to have the same vertex efficiency and edge efficiency as LWA* but it has a lower computational cost and memory cost.
Through simulation studies, we also show that the edge evaluation of weighted LEA* (wLEA*) is close to optimal compared to LazySP without the extra graph operations.  

The contributions of this paper are summarized as follows:
\begin{itemize}
    \item We propose the LEA* algorithm for the shortest path planning problem. LEA* uses an edge queue and imposes minimal modifications on A*, which results in a small overhead compared with previous lazy search algorithms.
    \item We prove the completeness, optimality, and optimal vertex efficiency of LEA*. 
    \item We show that the edge efficiency of wLEA* with an inflated heuristic is similar to LazySP, thus wLEA* has near-optimal edge efficiency.
    \item We test LEA* on various randomly generated environments and perform a thorough comparison with previous algorithms, demonstrating the efficiency of LEA*.
\end{itemize}


\section{Related Works} \label{SecRelatedWorks}
Graphs offer a powerful abstraction tool for robot path and motion planning.
Graphs can be divided into explicit graphs and implicit graphs.
When combined with sampling-based methods, algorithms such as PRM \cite{kavraki1996probabilistic}, PRM* \cite{Karaman2011Sampling}, and BIT* \cite{gammell2020batch} solve planning problems in very high-dimensional spaces.
They construct a graph of robot configurations and find the shortest path by exploiting this graph.
These algorithms use explicit graphs.
On the other hand, implicit graphs are defined implicitly using state lattice or motion primitives \cite{Likhachev2009Planning, Pivtoraiko2009Differentially, Liu2017Search}. By constructing motion primitives offline, they are beneficial in dealing with kinematic constraints, nonholonomic constraints, and differential constraints. 
Planning problems for ground vehicles and micro aerial vehicles are studied in \cite{Pivtoraiko2009Differentially} and \cite{Liu2017Search}, respectively. 

The graph search algorithms studied in this paper can be used with both explicit graphs and implicit graphs.
The A* search algorithm is a popular search algorithm that is vertex optimal \cite{Hart1968A}. 
Recent studies show that A* is not edge optimal \cite{Dellin2016A}. By employing a lazy approach, the number of edge evaluations can be reduced \cite{bohlin2000path, Cohen2015Planning, Hauser2015Lazy}. 
For instance, LWA* \cite{Cohen2015Planning} uses a one-step lookahead to postpone edge evaluation. 
It allows duplicate vertices in the vertex queue. A valid vertex with an estimated cost that is poped from the vertex queue is inserted into the queue again after edge evaluation.
LazySP \cite{Dellin2016A} uses an infinite-step lookahead and is shown to be edge optimal (evaluating the minimum number of edges). 
As the lookahead steps increase, the number of edge evaluations required to find the shortest path decrease, while the additional graph operation increase. 
Therefore, LRA* \cite{Mandalika2018Lazy} interpolates between LWA* and LazySP and uses a constant lookahead in the interval $[1, \infty]$.
A modified version of LWA* is also presented in \cite{Dellin2016A} by using both a vertex queue and an edge queue.
Compared to LWA*, our proposed LEA* algorithm shows that the vertex queue is unnecessary. By only maintaining an edge queue, LEA* achieves the same vertex efficiency and edge efficiency of LWA*. 
Thus, LEA* has a lower computational cost and possibly a lower memory usage. The advantage is more evident with large graphs as shown in the evaluation section.
The weighted LEA* (wLEA*) has similar edge efficiency as LazySP without the extra graph operation needed by LazySP.

In \cite{Mandalika2019Generalized}, the generalized lazy search algorithm (GLS) is proposed that unifies LWA*, LazySP, and LRA*. It also uses prior information of the planning environment to accelerate the search. 
A lazy incremental algorithm dealing with dynamic changing graphs is also introduced in \cite{Lim2022Lazy} by combining the idea of lazy search and lifelong planning \cite{Koenig2004Lifelong}. 

\section{Problem formulation} \label{SecProblemformulation}

We consider the problem of finding the shortest path between a starting vertex $v_s$ and a goal vertex $v_g$ on a graph $G = (V, E)$, where $V$ is the vertex set and $E$ is the edge set of the graph. 
Each edge $e$ has a cost $w(e)$ and the edge cost set is denoted as $W$.  

We consider the case when the edge cost set $W$ is unknown at the beginning of the graph search. 
This setting has practical advantages for robot path/motion planning problems.
Evaluating all edges before the graph search is unnecessary since only part of the edges will most likely be visited during the graph search to find the shortest path.  
Path planning for robots involves two main steps: constructing the graph and graph search.
Since edge evaluation (i.e., checking if the robot collides with the environment when moving along an edge) is expensive, evaluating only edges that are visited during the search saves the total planning time. 
Another advantage of this problem setting is that the graph $G$ is environment-independent. 
Since we do not perform collision checking when building the graph, $G$ can be used with different environments with different obstacles.

We assume that an admissible and consistent heuristic of the edge cost is available, which is easier to compute compared to the true cost. 
Specifically, we consider the special case where the estimated edge cost $\hat{w}$ is equal to the true cost $w$ without considering obstacles, and
\begin{equation}
    w(e) = \begin{cases} \hat{w}(e),  \ \quad \text{if $e$ is not in collision}, \\ 
    \enspace \infty,  \ \ \ \quad \text{if $e$ is in collision},
    \end{cases}
\end{equation}
where $\hat{w}(e)$ is the length of the edge assuming it is collision-free.

\section{The LEA* Algorithm} \label{LEA*}
In this section, we provide a detailed description of the proposed lazy edge-based A* (LEA*) algorithm.
LEA* uses the idea of lazy search and the tree expansion is ordered using an edge queue.
We design LEA* to have minimum changes compared to the original A* algorithm, and have as little computational overhead as possible.

Similar to A*, we define the \textit{cost-to-come} and the heuristic \textit{cost-to-go} of the vertex $v$ as $g(v)$ and $h(v)$, which are the actual cost from $v_s$ to $v$ given the current search tree and the estimated cost from $v$ to $v_g$, respectively. Here, $h(\cdot)$ is an admissible and consistent heuristic.
The total estimated cost of the path passing through $v$ is given by 
\begin{equation}
    f(v) = g(v) + h(v).
    \label{f_v}
\end{equation}
An edge is given by $e = (\underline{v}, \overline{v})$, where $\underline{v}$ is the source vertex and $\overline{v}$ is the target vertex.
We define the \textit{total estimated cost of edge} $e$ as
\begin{equation}
    f(e) = g(\underline{v}) + \hat{w}(e) + h(\overline{v}),
    \label{f_e}
\end{equation}
which is the total estimated cost of the path from $v_s$ to $v_g$ passing through $e$.

\begin{figure*}[t]
 \centering
   \begin{tabular}{@{}ccc@{}}
   \begin{minipage}{.26\textwidth}
    \includegraphics[width=\textwidth]{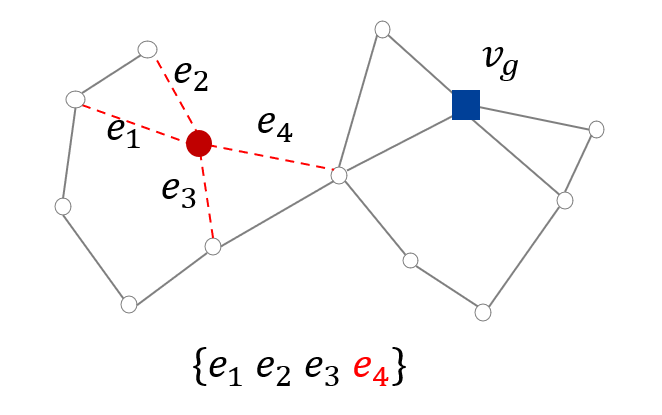}
     \captionof*{figure}{(a)}
   \end{minipage} &
    \begin{minipage}{.26\textwidth}
    \includegraphics[width=\textwidth]{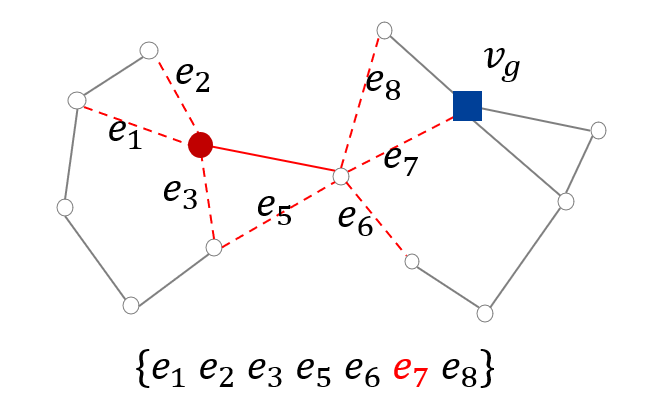}
    \captionof*{figure}{(b)}
   \end{minipage} &
    \begin{minipage}{.26\textwidth}
    \includegraphics[width=\textwidth]{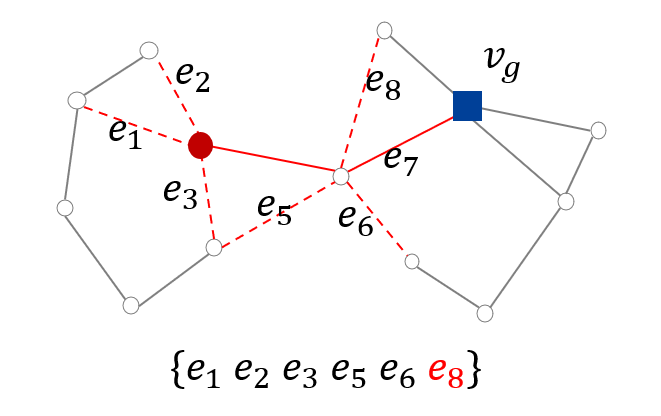}
    \captionof*{figure}{(c)}
   \end{minipage}\\
   \begin{minipage}{.26\textwidth}
    \includegraphics[width=\textwidth]{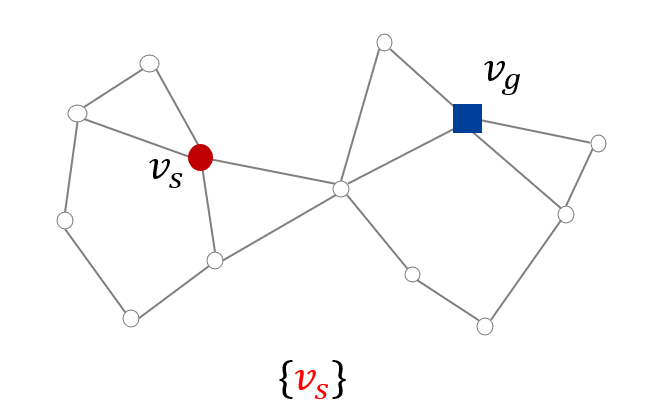}
    \captionof*{figure}{(d)}
   \end{minipage} &
   \begin{minipage}{.26\textwidth}
    \includegraphics[width=\textwidth]{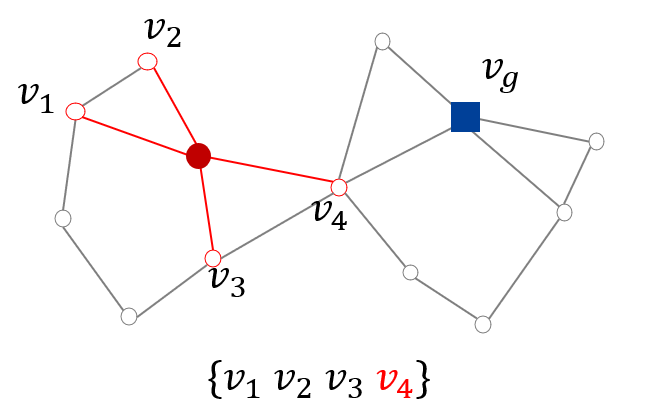}
    \captionof*{figure}{(e)}
   \end{minipage} &
    \begin{minipage}{.26\textwidth}
    \includegraphics[width=\textwidth]{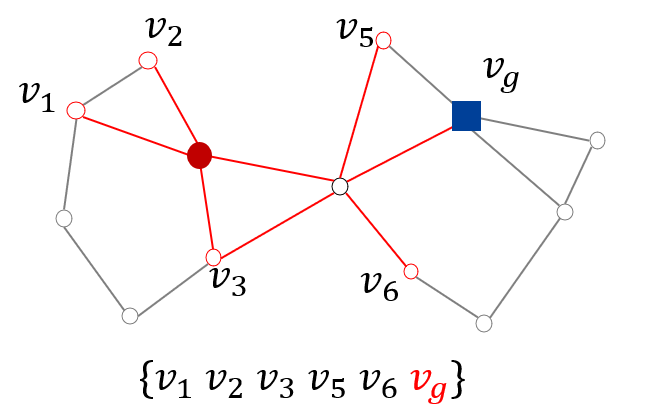}
    \captionof*{figure}{(f)}
   \end{minipage}
  \end{tabular}
    \caption{Illustration of the steps of LEA* ((a)-(c)) and A* ((d)-(f)). The set below each figure is the current edge queue or vertex queue. Dashed red lines are edges that are lazily evaluated. Solid red lines are evaluated edges. In (a), four edges are added to the edge queue, and $e_4$ is selected for evaluation. In (b), $e_4$ is evaluated, the next edges are added to the edge queue, and $e_7$ is selected. By using lazy evaluation and an edge queue, LEA* only evaluated two edges while A* evaluated eight edges.}
  \label{fig:LEA*illustration}
\end{figure*}

\IncMargin{.5em}
\begin{algorithm}
\caption{LEA*}
\label{alg:LEA*}
$Q_E \leftarrow \{(v_s, v_i)|v_i \in \mathsf{Succ}(v_s) \}$\;
\While{$Q_E \neq \emptyset$}
{   
    $(\underline{v}, \overline{v}), key \leftarrow Q_E.\mathsf{Pop}$\;
    \If{$g(v_g) \leq key$}
    {
        $\mathsf{return}$
    }
    $w \leftarrow \mathsf{Evaluate}((\underline{v}, \overline{v}))$\;
    \If{$w < \infty$}
    {
        $g_\mathrm{new} \leftarrow g(\underline{v}) + w$\;
        \If{$g_\mathrm{new}<g(\overline{v})$}
        {
            $g(\overline{v}) \leftarrow g_\mathrm{new}$\;
            $\overline{v}.\mathrm{parent} \leftarrow \underline{v}$\;
            $Q_E \leftarrow Q_E \cup \{(\overline{v}, v_i)|v_i \in \mathsf{Succ}(\overline{v}) \}$\;        
        }
    }
}
\end{algorithm}
\DecMargin{.5em}

The LEA* algorithm is given in Algorithm \ref{alg:LEA*}.
In Line~1, the edges from the starting vertex $v_s$ to its successors $v_i$ are added to the edge queue $Q_E$.
The edges in $Q_E$ are ordered by their $f$-values given by (\ref{f_e}).
In Line 3, the best edge $e=(\underline{v}, \overline{v})$, which has the smallest $key$, is popped from $Q_E$. 
Here $key$ is the $f$-value of $(\underline{v}, \overline{v})$ and is used in the termination condition in Line 4.
If the cost-to-come of goal vertex $g(v_g)$ is less than or equal to $key$, we have found the shortest path.
The true edge cost $w$ is computed in Line 6.
If $(\underline{v},\overline{v})$ is in collision, $w = \infty$. Otherwise, $w$ is the distance between $\underline{v}$ and $\overline{v}$.
Provided that $w < \infty$, the new cost-to-come of $\overline{v}$, $g_\mathrm{new}$, is computed in Line 8.
If $g_\mathrm{new}$ is better than the current cost-to-come $g(\overline{v})$, $g(\overline{v})$ is updated. The parent of $\overline{v}$ is also updated (Line 11).
Finally, the edges from $\overline{v}$ to its successors $v_i$ are added to the edge queue $Q_E$ in Line 12.
After the algorithm exits, we can get the shortest path from $v_s$ to $v_g$ using the parent information.

\IncMargin{.5em}
\begin{algorithm}
\caption{A*}
\label{alg:A*}
$Q_V \leftarrow \{v_s\}$\;
\While{$Q_V \neq \emptyset$}
{   
    $v_c \leftarrow Q_V.\mathsf{Pop}$\;
    \If{$v_c = v_g$}
    {
        $\mathsf{return}$
    }
    \ForEach{$v_i \in \mathsf{Succ}(v_c)$}
    {
        $w \leftarrow \mathsf{Evaluate}((v_c, v_i))$\;
        \If{$w < \infty$}
        {
            $g_\mathrm{new} \leftarrow g(v_c) + w$\;
            \If{$g_\mathrm{new}<g(v_i)$}
            {
                $g(v_i) \leftarrow g_\mathrm{new}$\;   
                $v_i.\mathrm{parent} \leftarrow v_c$\;
                $Q_V \leftarrow Q_V \cup \{v_i\}$\;    
                }
        }
    }  
}
\end{algorithm}
\DecMargin{.5em}

A simple illustration of the LEA* and A* algorithms is given in Figure~\ref{fig:LEA*illustration}.
Figures~\ref{fig:LEA*illustration}(a)-(c) are the steps of LEA* and Figures~\ref{fig:LEA*illustration}(d)-(f) are the steps of A*.
At the beginning of LEA*, four edges are added to the edge queue as shown in Figure~\ref{fig:LEA*illustration}(a). 
Note that these edges are not evaluated, instead, they are lazily evaluated using the estimated cost $\hat{w}$, and $\hat{w}$ is used to compute their $f$-values according to (\ref{f_e}). 
The best edge $e_4$ is selected for edge evaluation. 
In Figure~\ref{fig:LEA*illustration}(b), $e_4$ is evaluated and the next edges are added to the queue. 
Then, the best edge $e_7$ in the current queue is selected for evaluation.
For A*, all outgoing edges are evaluated when expanding a vertex.
In this example, by using lazy evaluation and an edge queue, LEA* only evaluated two edges to find the solution, while A* evaluated eight edges.

\section{Algorithmic Analysis}
In this section, we analyze the completeness, optimality, and vertex efficiency of LEA*.
\subsection{Completeness of LEA*}
\begin{theorem}
If at least one solution exists for the graph search problem, LEA* will return a solution. Otherwise, will return that no solution exists.
\end{theorem}
\begin{proof}
We first show that the algorithm terminates with $Q_E = \emptyset$ and $g(v_g) = \infty$ when no solution exists.
All edges have positive edge costs. The new edge $(\overline{v}, v_i)$ is added to $Q_E$ in Line 12, Algorithm~\ref{alg:LEA*}, only when we find a better path to $\overline{v}$ (Line 9). Since the cost-to-come to every vertex is decreasing and lower bounded, Line 12 will only run a finite number of iterations. Thus, $Q_E = \emptyset$ after executing Line 3 a finite number of iterations. Therefore, LEA* will terminate with $Q_E = \emptyset$ and $g(v_g) = \infty$ if no solution exists.

When the problem has a solution, LEA* terminates when $g(v_g) \leq key = \mathrm{min}_{e \in Q_E} f(e)$ or $Q_E = \emptyset$.
If the algorithm terminated with $g(v_g) \leq \mathrm{min}_{e \in Q_E} f(e)$, we have $g(v_g) < \infty$, which implies a solution has been found.

Let $(e_0,e_1, \dots, e_{n-1})$ (equivalently, $(v_s, v_1, v_2, \dots, v_g)$) be a solution path. 
Initially, $e_0 \in Q_E$. After evaluating $e_0$, $e_1$ is added to $Q_E$. Similarly after evaluating $e_1$, $e_2$ is added to $Q_E$.
If the algorithm terminated with $Q_E = \emptyset$, it must have evaluated $e_0,e_1, \dots, e_{n-1}$. Thus, the algorithm has found this solution and the returned solution is at least as good as $(e_0,e_1, \dots, e_{n-1})$.  
\end{proof}


\subsection{Optimality of LEA*}
\begin{theorem}
If at least one solution exists for the graph search problem, LEA* finds the minimum cost solution.
\end{theorem}
\begin{proof}
Let $\tau^* = (v_s, v_1^*, v_2^*, \dots, v_{n-1}^*, v_g)$ (equivalently $(e_0^*, e_1^*, \dots, e_{n-2}^*, e_{n-1}^*)$) be an optimal solution path with path cost $c^*$. 
Let $\tau = (v_s, v_1, v_2, \dots, v_{n-1}, v_g)$ (equivalently $(e_0, e_1, \dots, e_{n-2}, e_{n-1})$) be the solution path returned by LEA*. The cost of $\tau$ is $c$. 
We need to show that $c$ is equal to $c^*$ to prove optimality.

To reach a contradiction, let us assume $c^* < c$.
Note that $\max\{f(e_0^*), f(e_1^*), \dots, f(e_{n-2}^*), f(e_{n-1}^*)\} \leq c^*$. 
Also,  $g(v_g) = c \leq \mathrm{min}_{e \in Q_E} f(e)$ holds when LEA* terminates.
In the beginning of LEA*, $e_0^* \in Q_E$. Therefore, $e_0^*$ must have been evaluated before $c \leq \mathrm{min}_{e \in Q_E} f(e)$ is true. After evaluating $e_0^*$, $v_1^*$ is added to the expansion tree, and $e_1^*$ is added to $Q_E$. 
By repeating this analysis, $e_1^*, \dots, e_{n-2}^*, e_{n-1}^*$ must all have been evaluated before $c \leq \mathrm{min}_{e \in Q_E} f(e)$ is true.
After evaluating $e_1^*, \dots, e_{n-1}^*$, we have found a path to $v_g$ and $f(v_g) \leq c^*$. Since the $f(v_g)$ is nonincreasing, the algorithm will never return a path with cost $c>c^*$. Therefore $c=c^*$.
\end{proof}

\subsection{Optimal Vertex Efficiency}
\begin{theorem}
LEA* has the same vertex efficiency as A*. Furthermore, the expanded vertex set of LEA* is a subset of the expanded vertex set of A*.
\end{theorem}
\begin{proof}
For LEA*, we call evaluating any outgoing edge of $v$, expanding $v$.
A* expands $v$ by evaluating all outgoing edges of $v$. Vertex $v$ is expanded in LEA* if a subset of its outgoing edges is evaluated.

Let $\tau^* = (e_0^*, e_1^*, \dots, e_{n-2}^*, e_{n-1}^*)$ be the path found by A* and LEA*, and let $c^*$ be the cost of $\tau^*$.
Let $T=(V,E)$ and $T^e = (V^e, E^e)$ be the expansion tree grown by A* and LEA*, respectively.
Let $W^v$ and $W^e$ be the set of vertices expanded by A* and LEA*, respectively.
By showing $W^e \subseteq W^v$, we prove the optimally efficient search of 
LEA*.
Note that if $W^e \subseteq W^v$, we have $V^e \subseteq V$. 

Now assume $W^e \not\subseteq W^v$. Then, there exists $v$, such that $v \notin W^v$, $v \in W^e$, $v \in V^e$ and $v \in V$. 
To show that this is true, we start with $v_0 \in W^e$, and $v_0 \notin W^v$. Then, $v_0 \in V^e$ (vertex must be in the tree for it to be expanded). If $v_0 \notin V$, we find $v_1$, which is the parent of $v$ in $T^e$. Then, we have $v_1 \in V^e$, $v_1 \in W^e$, and $v_1 \notin W^v$. If $v_1 \notin V$, we repeat this process by finding its parent vertex in $V^e$, and one of these parents $v_i$ must satisfy $v_i \in V$ since $V$ and $V^e$ share the same root vertex.  

From $v \notin W^v$, we have $c^* \leq f(v) = g(v) + h(v)$.
From $v \in W^e$, we have that there exists $v_i \in \mathsf{Succ}(v)$, such that $f((v,v_i)) = g(v)+\hat{w}((v,v_i)) +h(v_i) \leq c^*$. Note that $f(v) \leq f((v,v_i))$. Then, we have
\begin{equation}
    c^* \leq f(v) \leq f((v,v_i)) \leq c^*,
\end{equation}
which only holds when $f(v) = f((v,v_i)) = c^*$.
Note that
\begin{equation}
    \max\{f(e_0^*), f(e_1^*), \dots, f(e_{n-2}^*), f(e_{n-1}^*)\} \leq c^*. \label{eq1}
\end{equation}
Using (\ref{eq1}), in order for $(v, v_i)$ to be evaluated by LEA*, there exist $e_i^* = (\underline{v}, \overline{v}) \in \tau^*$, such that $f(e_i^*) = c^*$ and $g(v) < g(\underline{v})$. Otherwise, all $e_i^* \in \tau^*$ have a higher priority than $(v, v_i)$ and $(v, v_i)$ will not be evaluated. 
Using $f(v) = c^* $ and $g(v) < g(\underline{v}) \leq g(v_g)$, $v$ has a higher priority than $v_g$. Therefore, $v$ must be expanded by A* before $v_g$ is selected from the vertex queue. Therefore, $v \in W^v$, contradicting $v \notin W^v$.
\end{proof}

\section{Experimental Results} \label{SecExperiment}
In this section, we compare LEA* with A*, LWA*, LazySP, and LRA*. Our code is available online\footnote{\texttt{\url{https://github.com/dongliangCH/LEAstar}}}. 
The lookahead parameter for LRA* is set to 4 in our implementations. 
Two problems are studied: a planning problem in 2D environments and planning for a 7DOF manipulator. 
\subsection{2D Planning Problem}
\begin{figure}[htb]
    \centering
    \includegraphics[width=0.9\columnwidth]{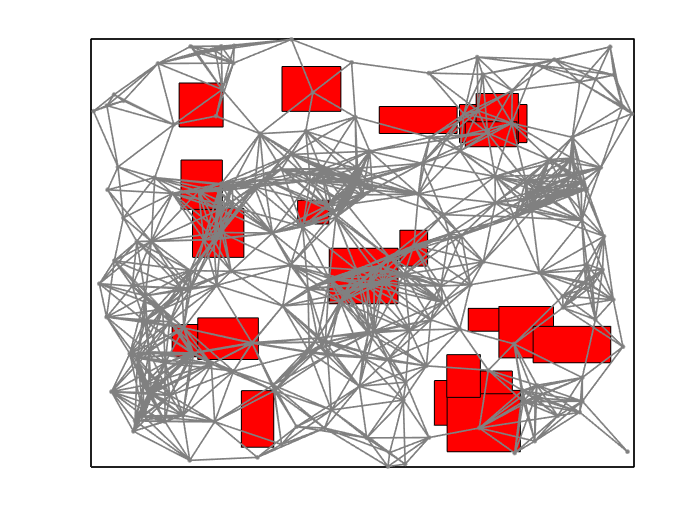}
    \caption{2D planning problem. Obstacles are sampled randomly. The graph is constructed without considering the presence of obstacles.}
    \label{2Dexample}
\end{figure}

The 2D environment is filled with randomly generated obstacles.
An example of the planning environment is given in Figure \ref{2Dexample}.
The 2D environment is divided into sparse, medium, and cluttered environments, each with 8, 16, and 24 obstacles, respectively.
Each obstacle is a rectangle, and its width and height are uniformly sampled from an interval.
The location of each obstacle is also uniformly sampled inside the boundary of the environment.

\begin{figure}[htb]
    \centering
    \begin{subfigure}[b]{0.49\columnwidth}
         \centering
         \includegraphics[width=\columnwidth]{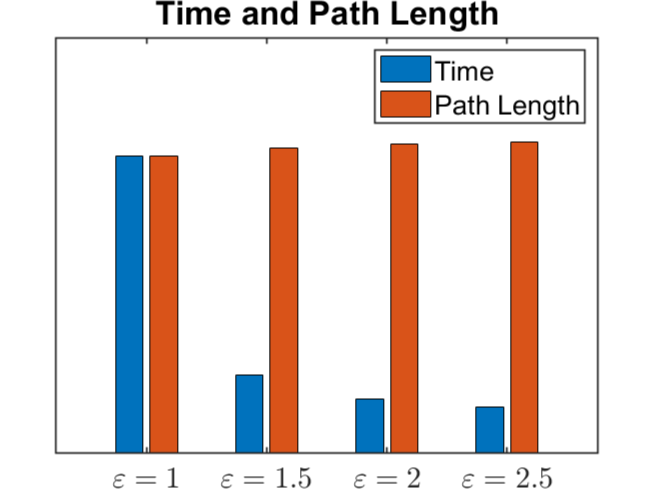}
         \caption{}
     \end{subfigure}
         \centering
    \begin{subfigure}[b]{0.49\columnwidth}
         \centering
         \includegraphics[width=\columnwidth]{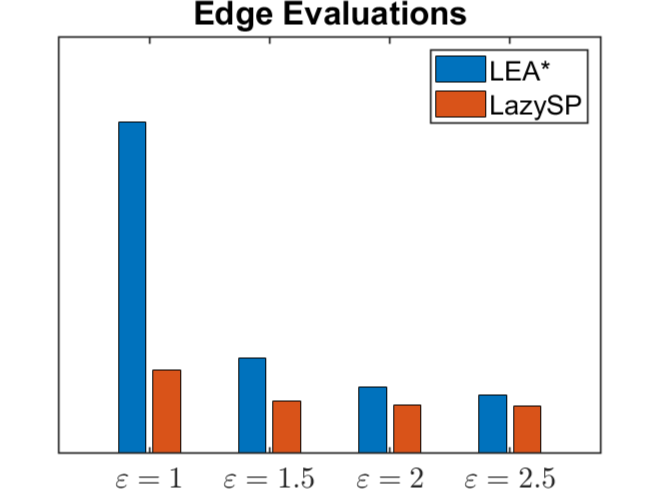}
         \caption{}
     \end{subfigure}
        \caption{(a) The trend of the planning time of LEA* and solution path length with different $\varepsilon$. The average planning time decreases $\sim85\%$ while the path length only increases $\sim5\%$. (b) The number of edge evaluations of LazySP and LEA* for different $\varepsilon$. The detailed results for $\varepsilon = 1$ and $\varepsilon = 2$ are given in Table~\ref{tab:2Dedge}. The gap between LazySP and LEA* becomes smaller as $\varepsilon$ increases.}
        \label{2Dresults}
\end{figure}

We also consider different graph sizes from small graphs to large graphs.
The graph size is indicated by the number of vertices $N$ in the graph. 
Specificly, $N = 200$, $N = 1,000$, $N = 5,000$, $N = 10,000$, $N = 20,000$ are considered.
Following \cite{Karaman2011Sampling}, the connecting radius of the graph is a function of $N$.  
The graph construction step follows the lazy PRM algorithm \cite{bohlin2000path}.
No collision check is performed when sampling vertices or adding edges.
As has been discussed in Section \ref{SecProblemformulation}, this will save planning time since evaluating all edges is unnecessary. 
In addition, after the graph is constructed, the same graph may be used with different environments.  
We sample 10 environments for each environment category (sparse, medium, cluttered).
For each environment and graph combination, 50 start-goal vertices are sampled as queries of the shortest path problem. The total number of planning problems for the 2D environment is therefore $7,500 = 3 \times 5 \times 10 \times 50$. 

The results are summarized in Tables~\ref{tab:2Dtime} and \ref{tab:2Dedge}.
The shortest paths returned by all five algorithms have the same path length for each tested case.   
The planning time results are given in Table~\ref{tab:2Dtime}. 
LEA* uses the least amount of time to find the same solution compared to other algorithms.
LWA* and LEA* have similar performance. 
The improvement of LEA* over LWA* is small but consistent, and the improvement becomes clearer with larger graphs since LWA* needs to maintain both vertex queue and edge queue.

Bounded suboptimal solutions can be easily obtained by inflating the cost-to-go heuristic $h(\cdot)$ with a factor $\varepsilon$. In our implementation, the inner loop of LazySP and LRA* use A* for lazy search. The values of $\varepsilon = \{1,1.5, 2, 2.5\}$ were tested.
All algorithms find the optimal path when $\varepsilon = 1$.
The results for $\varepsilon = 2$ are also given.
As we can see, the planning time decreased for all algorithms.
The trend for the planning time of LEA* and solution path length is given in Figure~\ref{2Dresults}(a).

The number of evaluated edges is given in Table~\ref{tab:2Dedge}.
Since LazySP is edge optimal, it evaluates the minimum number of edges.
However, as it requires extra graph operations such as vertex expansions and finding the current best path, the overall planning time is greater than LEA*.
Using an inflated heuristic $\varepsilon = 2$, the average number of edge evaluations are decreased for all algorithms.
The gap between LazySP and LEA* also becomes smaller.
The trend for edge evaluations with increasing inflation factor is given in Figure~\ref{2Dresults}(b).

\begin{table*}
\caption{Planning time of the 2D planning problem}\label{tab:2Dtime}
\centering
\begin{tabular}{L{1.8cm}|L{0.8cm}|L{0.8cm}|L{0.8cm}|L{0.8cm}|L{0.8cm}|L{0.8cm}|L{0.8cm}|L{0.8cm}|L{0.8cm}|L{0.8cm}|L{0.8cm}|L{0.8cm}}
\toprule
\hline
\multirow{2}{1.8cm} {\centering Avg. Time (s)} & \multicolumn{3}{c|}{N = 1000}   & \multicolumn{3}{c|}{N = 5000} & \multicolumn{3}{c|}{N = 10000} & \multicolumn{3}{c}{N = 20000} \\
\cline{2-13}  &  o = 8  & o = 18  &  o = 28 &  o = 8  & o = 18  &  o = 28  & o = 8  & o = 18  &  o = 28 &  o = 8  & o = 18  &  o = 28 \\             
\hline                                                                           
A* ($\varepsilon=1$)       & 0.0871   & 0.0998   & 0.1036  & 0.2270  & 0.2397   & 0.2761   & 0.3500  & 0.3913 & 0.4655  & 0.5358  & 0.5952 & 0.6846 \\
LWA* ($\varepsilon=1$)     & 0.0118   & 0.0150   & 0.0167  & 0.0424  & 0.0495    & 0.0618   & 0.0781  & 0.0895 & 0.1152  & 0.1436  & 0.1676 & 0.2061\\
LEA* ($\varepsilon=1$)     & \textbf{0.0109}   & \textbf{0.0137}   & \textbf{0.0152}  & \textbf{0.0378}  & \textbf{0.0428}   & \textbf{0.0530}   & \textbf{0.0682}  & \textbf{0.0771} & \textbf{0.0965}  & \textbf{0.1204}  & \textbf{0.1403} & \textbf{0.1671} \\
LazySP ($\varepsilon=1$)   & 0.0236   & 0.0690   & 0.1104  & 0.2030  & 0.6452   & 1.2149   & 0.5167  & 1.4903 & 3.0549  & 1.7829  & 3.7850 & 5.1193 \\
LRA* ($\varepsilon=1$)     & 0.0393   & 0.0799   & 0.1182  & 0.4536  & 0.8561   & 1.5409   & 1.4263  & 2.3989 & 4.3414  & 4.4443  & 5.6916 & 7.3189 \\ 
\hline
A* ($\varepsilon=2$)      & 0.0198   & 0.0246   & 0.0269  & 0.0278  & 0.0329   & 0.0542   & 0.0329  & 0.0409 & 0.1199  & 0.0739  & 0.0654 & 0.1078 \\
LWA* ($\varepsilon=2$)    & 0.0031   & \textbf{0.0040}   & 0.0053  & \textbf{0.0050}  & 0.0079   & 0.0142   & 0.0081  & 0.0099 & 0.0353  & 0.0239  & 0.0222 & 0.0402 \\
LEA* ($\varepsilon=2$)    & \textbf{0.0028}   & \textbf{0.0040}   & \textbf{0.0049}  & 0.0051  & \textbf{0.0070}   & \textbf{0.0122}   & \textbf{0.0071}  & \textbf{0.0093} & \textbf{0.0264}  & \textbf{0.0182}  & \textbf{0.0179} & \textbf{0.0322} \\
LazySP ($\varepsilon=2$)  & 0.0063   & 0.0175   & 0.0238  & 0.0218  & 0.0622   & 0.1771   & 0.0438  & 0.1101 & 0.4731  & 0.2760  & 0.3312 & 0.8808 \\
LRA* ($\varepsilon=2$)    & 0.0070   & 0.0157   & 0.0227  & 0.0235  & 0.0573   & 0.1935   & 0.0470  & 0.0995 & 0.6607  & 0.5203  & 0.4129 & 1.0473 \\ 
\bottomrule
\end{tabular}
\end{table*}

\begin{table*}
\caption{Number of edge evaluations of the 2D planning problem}\label{tab:2Dedge}
\centering
\begin{tabular}{L{1.8cm}|L{0.8cm}|L{0.8cm}|L{0.8cm}|L{0.8cm}|L{0.8cm}|L{0.8cm}|L{0.8cm}|L{0.8cm}|L{0.8cm}|L{0.8cm}|L{0.8cm}|L{0.8cm}}
\toprule
\hline
\multirow{2}{1.8cm} {\centering Edge num.} & \multicolumn{3}{c|}{N = 1000}   & \multicolumn{3}{c|}{N = 5000} & \multicolumn{3}{c|}{N = 10000} & \multicolumn{3}{c}{N = 20000} \\
\cline{2-13}  &  o = 8  & o = 18  &  o = 28 &  o = 8  & o = 18  &  o = 28  & o = 8  & o = 18  &  o = 28 &  o = 8  & o = 18  &  o = 28 \\  
\hline
A* ($\varepsilon=1$)      & 900.3   & 1008.5  & 1045   & 4049.3 & 4265.1   & 4913.8   & 7718.2  & 8279.2 & 9980.1  & 14434   & 16138  & 18666 \\
LWA* ($\varepsilon=1$)    & 49.39   & 64.35   & 74.89  & 164.7  & 196.58   & 252.27   & 294.32  & 318.44 & 410.43  & 501.89  & 588.78 & 720.81 \\
LEA* ($\varepsilon=1$)    & 49.39   & 64.35   & 74.89  & 164.7  & 196.58   & 252.27   & 294.32  & 318.44 & 410.43  & 501.89  & 588.78 & 720.81 \\
LazySP ($\varepsilon=1$)  & \textbf{15.87}   & \textbf{26.05}   & \textbf{33.53}  & \textbf{34.02} & \textbf{57.97}    & \textbf{82.06}    & \textbf{49.75}   & \textbf{80.26}  & \textbf{114.23}  & \textbf{76.98}   & \textbf{118.26} & \textbf{174.85} \\
LRA* ($\varepsilon=1$)    & 23.60   & 33.84   & 41.94  & 75.39  & 98.56    & 132.27   & 133     & 156.06 & 208.88  & 227.69  & 281.40 & 359.58 \\ 
\hline
A* ($\varepsilon=2$)      & 201.6   & 251.7   & 282.08 & 485.2  & 590.95  & 981.50  & 711.76 & 865.41 & 2510.3  & 1934.5 & 1781.1 & 3010.2 \\
LWA* ($\varepsilon=2$)    & 13.33   & 21.84   & 28.59  & 24.69  & 41.06   & 73.69   & 35.44  & 51.87  & 130.57  & 85.87  & 94.73  & 173.74 \\
LEA* ($\varepsilon=2$)    & 13.33   & 21.84   & 28.59  & 24.69  & 41.06   & 73.69   & 35.44  & 51.87  & 130.57  & 85.87  & 94.73  & 173.74 \\
LazySP ($\varepsilon=2$)  & \textbf{12.79}   & \textbf{19.08}   & \textbf{23.81}  & \textbf{23.74}  & \textbf{34.59}   & \textbf{48.15}   & \textbf{33.16}  & \textbf{43.62}  & \textbf{63.42}   & \textbf{46.71}  & \textbf{59.02}  & \textbf{86.78} \\
LRA* ($\varepsilon=2$)    & 12.82   & 19.20   & 24.03  & 23.76  & 34.89   & 50.07   & 33.27  & 44.09  & 68.01   & 50.12  & 60.88  & 94.36 \\ 
\bottomrule
\end{tabular}
\end{table*}

\begin{table*}
\caption{Planning time of the 7D planning problem}\label{tab:7Dtime}
\centering
\begin{tabular}{L{1.8cm}|L{0.8cm}|L{0.8cm}|L{0.8cm}|L{0.8cm}|L{0.8cm}|L{0.8cm}|L{0.8cm}|L{0.8cm}|L{0.8cm}}
\toprule
\hline
\multirow{2}{1.8cm} {\centering Avg. Time (s)} & \multicolumn{3}{c|}{N = 1000}   & \multicolumn{3}{c|}{N = 5000} & \multicolumn{3}{c}{N = 10000} \\
\cline{2-10}  &  o = 4  & o = 8  & o = 12 &  o = 4  & o = 8  &  o = 12  &  o = 4  & o = 8  &  o = 12 \\
\hline             
A* ($\varepsilon=1$)       & 0.9132   & 0.5835   & 0.5352  & 1.3899  & 1.5648   & 2.2098   & 3.8707  & 3.7700 & 3.0835  \\
LWA* ($\varepsilon=1$)     & 0.0952   & 0.0706   & 0.0696  & 0.0809  & 0.0866   & 0.1825   & 0.1657  & 0.1964 & 0.1916  \\
LEA* ($\varepsilon=1$)     & \textbf{0.0936}   & \textbf{0.0689}   & \textbf{0.0665}  & \textbf{0.0752}  & \textbf{0.0844}   & \textbf{0.1773}   & \textbf{0.1590}  & \textbf{0.1872} & \textbf{0.1802}  \\
LazySP ($\varepsilon=1$)   & 0.1419   & 0.1018   & 0.0743  & 0.0900  & 0.0808   & 0.8872   & 0.2299  & 0.9103 & 0.8298   \\
LRA* ($\varepsilon=1$)     & 0.1649   & 0.1720   & 0.0852  & 0.0955  & 0.0908   & 0.9601   & 0.2573  & 0.9951 & 0.8960   \\ 
\hline
A* ($\varepsilon=2$)      & 0.3139   & 0.2294   & 0.2466  & 0.3601  & 0.3448   & 0.4351   & 0.5266  & 0.4690 & 0.4680   \\
LWA* ($\varepsilon=2$)    & 0.0337   & 0.0280   & 0.0293  & 0.0188  & 0.0199   & 0.0331   & 0.0225  & 0.0246 & 0.0285   \\
LEA* ($\varepsilon=2$)    & \textbf{0.0322}   & \textbf{0.0276}   & \textbf{0.0290}  & \textbf{0.0183}  & \textbf{0.0193}   & \textbf{0.0325}   & \textbf{0.0210}  & \textbf{0.0233} & \textbf{0.0272}   \\
LazySP ($\varepsilon=2$)  & 0.0588   & 0.0376   & 0.0330  & 0.0259  & 0.0251   & 0.0788   & 0.0331  & 0.0482 & 0.0602   \\
LRA* ($\varepsilon=2$)    & 0.0842   & 0.0941   & 0.0344  & 0.0250  & 0.0252   & 0.0803   & 0.0343  & 0.0485 & 0.1052   \\ 
\bottomrule
\end{tabular}
\end{table*}

\begin{table*}
\caption{Number of edge evaluations of the 7D planning problem}\label{tab:7Dedge}
\centering
\begin{tabular}{L{1.8cm}|L{0.8cm}|L{0.8cm}|L{0.8cm}|L{0.8cm}|L{0.8cm}|L{0.8cm}|L{0.8cm}|L{0.8cm}|L{0.8cm}}
\toprule
\hline
\multirow{2}{1.8cm} {\centering Edge num.} & \multicolumn{3}{c|}{N = 1000}   & \multicolumn{3}{c|}{N = 5000} & \multicolumn{3}{c}{N = 10000} \\
\cline{2-10}  &  o = 4  & o = 8  & o = 12 &  o = 4  & o = 8  &  o = 12  &  o = 4  & o = 8  &  o = 12 \\
\hline                                          
A* ($\varepsilon=1$)      & 418.36  & 309.49  & 266.02  & 688.60  & 784.49  & 1279.0   & 1846.9 & 2106.7 & 1834.1  \\
LWA* ($\varepsilon=1$)    & 60.20   & 49.21   & 44.94   & 46.67   & 51.91   & 149.08   & 92.06  & 143.94 & 152.44  \\
LEA* ($\varepsilon=1$)    & 60.20   & 49.21   & 44.94   & 46.67   & 51.91   & 149.08   & 92.06  & 143.94 & 152.44  \\
LazySP ($\varepsilon=1$)  & \textbf{17.49}   & \textbf{16.58}   & \textbf{13.05}   & \textbf{9.13}    & \textbf{8.25}    & \textbf{27.95}    & \textbf{11.34}  & \textbf{18.89}  & \textbf{22.85}   \\
LRA* ($\varepsilon=1$)    & 19.19   & 19.01   & 13.25   & 9.23    & 8.77    & 30.35    & 13.28  & 23.15  & 26.11   \\ 
\hline
A* ($\varepsilon=2$)      & 164.25  & 139.50  & 138.28 & 203.21 & 200.07   & 303.81 & 285.1  & 307.9 & 334.2   \\
LWA* ($\varepsilon=2$)    & 21.51   & 19.92   & 17.94  & 10.50  & 11.70   & 27.02   & 12.65  & 17.57 & 21.99   \\
LEA* ($\varepsilon=2$)    & 21.51   & 19.92   & 17.94  & 10.50  & 11.70   & 27.02   & 12.65  & 17.57 & 21.99   \\
LazySP ($\varepsilon=2$)  & \textbf{16.14}   & \textbf{16.23}   & \textbf{13.00}  & \textbf{8.95}   & \textbf{9.11}   & \textbf{21.53}    & \textbf{10.68}  & \textbf{15.17} & \textbf{17.99}   \\
LRA* ($\varepsilon=2$)    & 18.34   & 18.90   & 13.05  & \textbf{8.95}   & \textbf{9.11}   & 21.60    & 10.73  & 15.18 & 18.57   \\ 
\bottomrule
\end{tabular}
\end{table*}

\begin{figure}[htb]
    \centering
    \includegraphics[width=0.78\columnwidth]{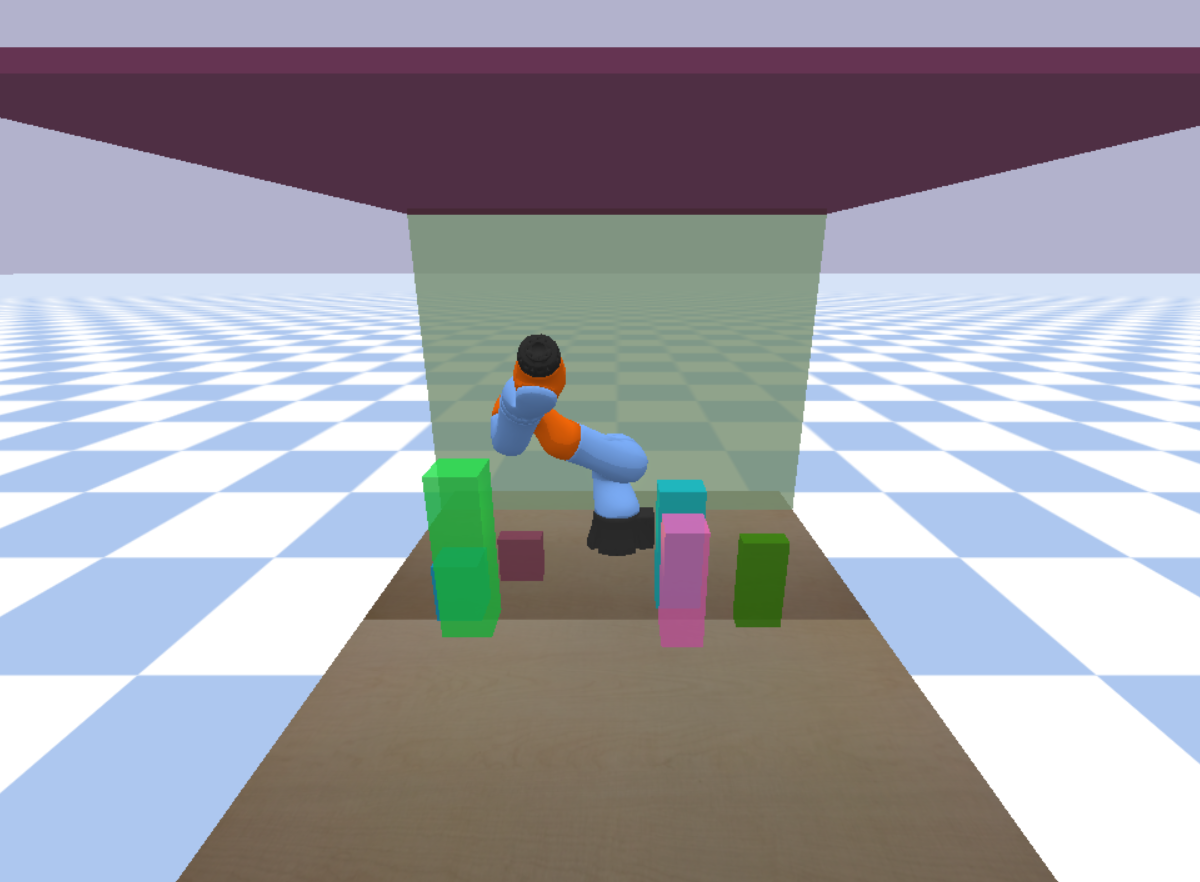}
    \caption{Manipulator tabletop example.}
    \label{7Dexample}
\end{figure}

\begin{figure}[htb]
    \centering
    \begin{subfigure}[b]{0.49\columnwidth}
         \centering
         \includegraphics[width=\columnwidth]{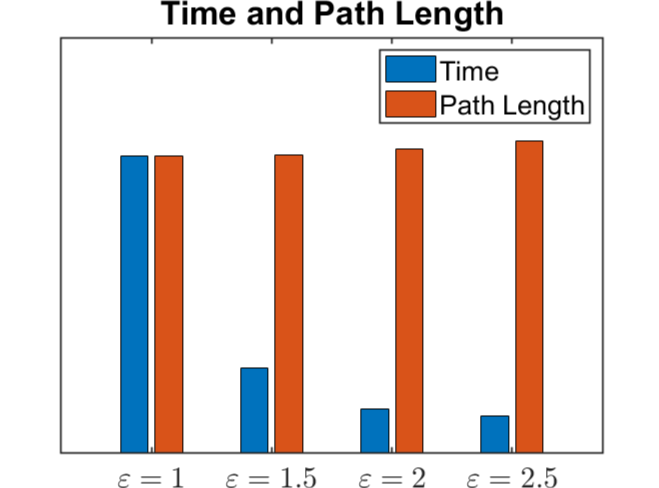}
         \caption{}
     \end{subfigure}
         \centering
    \begin{subfigure}[b]{0.49\columnwidth}
         \centering
         \includegraphics[width=\columnwidth]{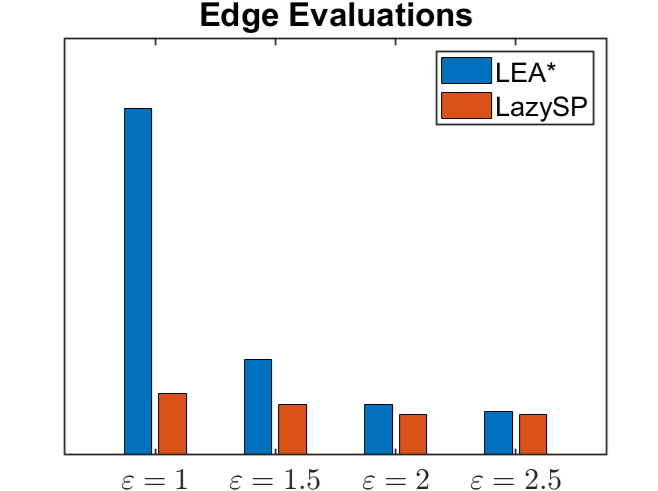}
         \caption{}
     \end{subfigure}
        \caption{Manipulator example. (a) The trend of the planning time of LEA* and solution path length with different $\varepsilon$. (b) The number of edge evaluations of LazySP and LEA* for different $\varepsilon$.}
        \label{7Dresults}
\end{figure}

\subsection{Planning for 7DOF Manipulator}
We consider the planning problem of a KUKA robot on a tabletop. The algorithms are implemented using the Pybullet simulator \cite{Coumans2021Pybullet}.
The planning environment is shown in Figure~\ref{7Dexample}.
Cubic obstacles are randomly generated on the tabletop.
The environment is divided into sparse, medium, and cluttered environments where each of them has 4, 8, and 12 obstacles, respectively.
We also set a wall and a ceiling to limit the operation space of the manipulator. 
The width, depth, and height of each cubic obstacle is uniformly sampled from its respective interval.
The location of the cubic obstacle is also sampled on the tabletop.
We consider small graphs, medium graphs, and large graphs with $N = 1,000$, $N = 5,000$, and $N = 10,000$, respectively. 
We sample ten environments for each environment category and sample 50 start-goal queries for each environment and graph combination. Therefore, the total number of planning problems is $4,500$. 

The results are summarized in Tables~\ref{tab:7Dtime} and \ref{tab:7Dedge}.   
The planning time results are given in Table~\ref{tab:7Dtime}. 
LEA* uses the least amount of time to find the same solution compared to other algorithms.
Bounded suboptimal solutions are obtained by using an inflation factor. 
The values $\varepsilon = \{1,1.5, 2, 2.5\}$ were tested.
The planning time decreases significantly for all algorithms.
The trend for the planning time of LEA* and solution path length is given in Figure~\ref{7Dresults}(a).
The number of evaluated edges is given in Table~\ref{tab:7Dedge}. The results are consistent with the 2D planning problem.
LazySP evaluates the minimum number of edges and is edge optimal.
The trend for edge evaluations with increasing inflation factor is given in Figure~\ref{7Dresults}(b). The gap between LazySP and LEA* becomes smaller as $\varepsilon$ increases.

\section{Conclusion}  \label{secConclusion}
In this work, we introduce the LEA* algorithm for robot path planning. 
LEA* is complete, optimal, with optimal vertex efficiency, and improved edge efficiency.
LEA* imposes only a few modifications to A*, making it easy to implement.
We show that the number of evaluated edges of weighted LEA* becomes close to the optimal value, while avoiding the computational overhead of LazySP and LRA*. 
We benchmark our algorithm with previous algorithms on various randomized environments.  
Our results show that LEA* and weighted LEA* are the fastest algorithms to find the plan in all tested examples. 
Future work includes using LEA* and motion primitives for motion planning of robots with nonholonomic and differential constraints.

\balance

\bibliographystyle{ieeetr}
\bibliography{references}

\begin{thebibliography}{10}

\bibitem{kavraki1996probabilistic}
L.~E. Kavraki, P.~Svestka, J.~C. Latombe, and M.~H. Overmars, ``Probabilistic
  roadmaps for path planning in high-dimensional configuration spaces,'' {\em
  IEEE Transactions on Robotics and Automation}, vol.~12, no.~4, pp.~566--580,
  1996.

\bibitem{bohlin2000path}
R.~Bohlin and L.~E. Kavraki, ``Path planning using lazy {PRM},'' in {\em
  Proceedings ICRA. Millennium Conference. IEEE International Conference on
  Robotics and Automation. Symposia Proceedings}, vol.~1, (San Francisco, CA),
  pp.~521--528, April 24-28, 2000.

\bibitem{gammell2020batch}
J.~D. Gammell, T.~D. Barfoot, and S.~S. Srinivasa, ``{Batch Informed Trees
  (BIT*)}: Informed asymptotically optimal anytime search,'' {\em The
  International Journal of Robotics Research}, vol.~39, no.~5, pp.~543--567,
  2020.

\bibitem{Hart1968A}
P.~E. Hart, N.~J. Nilsson, and B.~Raphael, ``A formal basis for the heuristic
  determination of minimum cost paths,'' {\em IEEE Transactions on Systems
  Science and Cybernetics}, vol.~4, no.~2, pp.~100--107, 1968.

\bibitem{LaValle2006Planning}
S.~M. LaValle, {\em Planning algorithms}.
\newblock Cambridge University Press, 2006.

\bibitem{Cohen2015Planning}
B.~Cohen, M.~Phillips, and M.~Likhachev, ``Planning single-arm manipulations
  with {N}-arm robots,'' in {\em Robotics: Science and Systems}, (Berkeley,
  CA), July 13-17, 2014.

\bibitem{Dellin2016A}
C.~Dellin and S.~Srinivasa, ``A unifying formalism for shortest path problems
  with expensive edge evaluations via lazy best-first search over paths with
  edge selectors,'' in {\em Proceedings of the International Conference on
  Automated Planning and Scheduling}, vol.~26, (London, UK), pp.~459--467, June
  12-17, 2016.

\bibitem{Mandalika2018Lazy}
A.~Mandalika, O.~Salzman, and S.~Srinivasa, ``Lazy receding horizon {A*} for
  efficient path planning in graphs with expensive-to-evaluate edges,'' in {\em
  Proceedings of the International Conference on Automated Planning and
  Scheduling}, vol.~28, (Delft, The Netherlands), pp.~476--484, June 24-29
  2018.

\bibitem{Karaman2011Sampling}
S.~Karaman and E.~Frazzoli, ``Sampling-based algorithms for optimal motion
  planning,'' {\em The International Journal of Robotics Research}, vol.~30,
  pp.~846--894, June 2011.

\bibitem{Likhachev2009Planning}
M.~Likhachev and D.~Ferguson, ``Planning long dynamically feasible maneuvers
  for autonomous vehicles,'' {\em The International Journal of Robotics
  Research}, vol.~28, no.~8, pp.~933--945, 2009.

\bibitem{Pivtoraiko2009Differentially}
M.~Pivtoraiko, R.~A. Knepper, and A.~Kelly, ``Differentially constrained mobile
  robot motion planning in state lattices,'' {\em Journal of Field Robotics},
  vol.~26, no.~3, pp.~308--333, 2009.

\bibitem{Liu2017Search}
S.~Liu, N.~Atanasov, K.~Mohta, and V.~Kumar, ``Search-based motion planning for
  quadrotors using linear quadratic minimum time control,'' in {\em IEEE/RSJ
  International Conference on Intelligent Robots and Systems (IROS)},
  (Vancouver, Canada), pp.~2872--2879, September 24–28, 2017.

\bibitem{Hauser2015Lazy}
K.~Hauser, ``Lazy collision checking in asymptotically-optimal motion
  planning,'' in {\em IEEE International Conference on Robotics and Automation
  (ICRA)}, (Seattle, WA), pp.~2951--2957, May 25-30, 2015.

\bibitem{Mandalika2019Generalized}
A.~Mandalika, S.~Choudhury, O.~Salzman, and S.~Srinivasa, ``Generalized lazy
  search for robot motion planning: Interleaving search and edge evaluation via
  event-based toggles,'' in {\em Proceedings of the International Conference on
  Automated Planning and Scheduling}, vol.~29, (Berkeley, CA), pp.~745--753,
  July 11-15, 2019.

\bibitem{Lim2022Lazy}
J.~Lim, S.~Srinivasa, and P.~Tsiotras, ``Lazy lifelong planning for efficient
  replanning in graphs with expensive edge evaluation,'' in {\em IEEE/RSJ
  International Conference on Intelligent Robots and Systems (IROS)}, (Kyoto,
  Japan), pp.~8778--8783, October 23-27, 2022.

\bibitem{Koenig2004Lifelong}
S.~Koenig, M.~Likhachev, and D.~Furcy, ``Lifelong planning {A*},'' {\em
  Artificial Intelligence}, vol.~155, no.~1, pp.~93--146, 2004.

\bibitem{Coumans2021Pybullet}
E.~Coumans and Y.~Bai, ``Pybullet, a python module for physics simulation for
  games, robotics and machine learning.'' \url{http://pybullet.org},
  2016--2021.

\end{thebibliography}

\addtolength{\textheight}{-12cm}   

\end{document}